\documentclass[11pt]{article}

\usepackage[utf8]{inputenc}
\usepackage[T1]{fontenc}

\usepackage[margin=1in]{geometry}
\usepackage{setspace}
\usepackage{microtype}

\usepackage[colorlinks=true,citecolor=blue,linkcolor=blue,urlcolor=blue]{hyperref}
\usepackage{url}
\usepackage[numbers]{natbib}

\usepackage{booktabs}
\usepackage{tabularx}

\usepackage{graphicx}

\usepackage{listings}
\usepackage{xcolor}

\usepackage{enumitem}

\usepackage{amsmath}

\title{\textbf{The Last Fingerprint: How Markdown Training Shapes LLM Prose}}

\author{
  E. M. Freeburg \\
  Independent Researcher \\
  \texttt{hqops@icloud.com}
}

\date{March 2026}

\begin{document}

\maketitle

\begin{abstract}
Large language models produce em dashes at varying rates, and the observation that some models ``overuse'' them has become one of the most widely discussed markers of AI-generated text. Yet no mechanistic account of this pattern exists, and the parallel observation that LLMs default to markdown-formatted output has never been connected to it. We propose that the em dash is \textit{markdown leaking into prose}---the smallest surviving unit of the structural orientation that LLMs acquire from markdown-saturated training corpora. We present a five-step genealogy connecting training data composition, structural internalization, the dual-register status of the em dash, and post-training amplification. We test this with a two-condition suppression experiment across twelve models from five providers (Anthropic, OpenAI, Meta, Google, DeepSeek): when models are instructed to avoid markdown formatting, overt features (headers, bullets, bold) are eliminated or nearly eliminated, but em dashes persist---except in Meta's Llama models, which produce none at all. Em dash frequency and suppression resistance vary from 0.0 per 1,000 words (Llama) to 9.1 (GPT-4.1 under suppression), functioning as a signature of the specific fine-tuning procedure applied. A three-condition suppression gradient shows that even explicit em dash prohibition fails to eliminate the artifact in some models, and a base-vs-instruct comparison confirms that the latent tendency exists pre-RLHF. These findings connect two previously isolated online discourses and reframe em dash frequency as a diagnostic of fine-tuning methodology rather than a stylistic defect.
\end{abstract}

\section{Introduction}
\label{sec:introduction}

Among the most widely discussed characteristics of text generated by large language models is the frequency of the em dash. Since the broad deployment of instruction-tuned models beginning in late 2022, users across writing communities, technology forums, and social media have identified em dash frequency as a marker of machine-generated prose---a ``tell'' that has become the subject of extensive online discussion, mockery, and ad hoc detection heuristics.

Separately, a parallel conversation has developed in developer and AI research communities: the observation that LLMs ``think in markdown.'' Models default to structured output---headers, bullet points, bold emphasis, numbered lists---even when such formatting is not requested. This tendency is widely attributed to the prevalence of markdown-formatted text in training corpora and is discussed primarily as a user experience or prompt engineering concern.

These two conversations describe the same phenomenon at different granularities, but they have not been connected. The writing communities observing em dash frequency do not frame their complaint in terms of training data composition or structural formatting. The developer communities analyzing markdown behavior do not extend their analysis to prose-level artifacts like punctuation patterns. A gap exists between the aesthetic observation (``AI overuses em dashes'') and the technical observation (``AI thinks in markdown''), and this paper fills it.

We propose that the em dash is \textbf{markdown leaking into prose}---the smallest surviving unit of the structural orientation that LLMs acquire from markdown-saturated training corpora. When models are constrained to produce natural prose rather than formatted output, the structural impulse is largely suppressed: headers, bullets, and bold emphasis are withheld. But the em dash survives in some models because it occupies a unique dual-register position: it is simultaneously valid prose punctuation and a structural marker. The instruction to ``write prose, not markdown'' passes over it because the em dash is already prose-legal.

Crucially, not all models exhibit this pattern. Our measurements show that Meta's Llama models produce zero em dashes across tens of thousands of words, while OpenAI and Anthropic models show elevated rates---with dramatic differences in how effectively suppression instructions reduce them. This variation is consistent with the em dash functioning not as a universal training artifact but as a \textbf{signature of the specific fine-tuning procedure applied}: different post-training methodologies amplify or suppress the latent tendency to different degrees.

\paragraph{Contributions.} This paper makes four contributions:

\begin{enumerate}[leftmargin=*]
\item A \textbf{five-step genealogy} for em dash production in LLM output, connecting training data composition, structural internalization, the dual-register status of the em dash, and post-training amplification into a single explanatory mechanism (Section~\ref{sec:genealogy}).

\item The \textbf{``last fingerprint'' framework}: a characterization of the em dash as the smallest unit of markdown-derived structural thinking that survives compression into prose, explaining why it persists under output constraints in models whose RLHF amplifies it (Section~\ref{sec:genealogy}).

\item \textbf{Empirical measurement} of em dash frequency and suppression resistance across twelve instruction-tuned LLMs from five providers ($\sim$240,000 words of generated prose), including a three-condition suppression gradient and a base-vs-instruct comparison, demonstrating that em dash behavior functions as a signature of fine-tuning methodology (Section~\ref{sec:empirical}).

\item A \textbf{documentation of two parallel discourses} about LLM writing behavior that have not been connected, and a demonstration that connecting them yields a unified explanation for both formatting-level and prose-level artifacts (Section~\ref{sec:discourses}).
\end{enumerate}

\section{Background and Related Work}
\label{sec:background}

\subsection{Markdown and LLM Training Data}

Markdown was published in 2004 by John Gruber, in collaboration with Aaron Swartz, as ``a text-to-HTML conversion tool for web writers''~\cite{gruber2004markdown}, formalizing conventions that writers were already employing informally in plain text. It has since become the dominant formatting layer for structured text on the internet, adopted by GitHub~\cite{github2024octoverse}, Stack Overflow, Reddit, Discord, Slack, and numerous documentation platforms. Two major standardization efforts---CommonMark~\cite{commonmark2021} and GitHub Flavored Markdown~\cite{gfm2024}---have formalized the syntax, while Pandoc~\cite{pandoc2024} provides the most comprehensive processing pipeline, including smart typography extensions that convert certain character sequences (notably \texttt{---} to em dash) during rendering.

The training data for large language models draws heavily from this markdown-saturated web. The Pile~\cite{gao2020pile}, an 800GB dataset widely used for language model training, includes GitHub repositories (where essentially all prose content is markdown), Stack Exchange (markdown-formatted), and a broad web crawl in which markdown-formatted content is disproportionately represented among high-quality, well-structured text. RedPajama~\cite{together2023redpajama} and RefinedWeb~\cite{penedo2023refinedweb} follow similar patterns. Audits of large web corpora~\cite{dodge2021documenting, longpre2023pretrainers} confirm that high-quality text sources are disproportionately drawn from platforms where markdown is the native formatting layer.

Precise estimates of the proportion of markdown-formatted text in LLM training corpora are not publicly available. However, the structural evidence is suggestive: the platforms that generate the highest-quality, most heavily structured text online---developer documentation, technical writing, academic and educational content---overwhelmingly use markdown. Training data curation processes that select for text quality therefore select, implicitly, for markdown-formatted text. The structural conventions of markdown---headings for hierarchical organization, dashes for boundaries and separators, asterisks for emphasis, lists for discrete enumeration---are not peripheral features of the training distribution. They are the organizational grammar of the highest-quality text in the corpus.

\subsection{RLHF and Output Style}

Reinforcement learning from human feedback (RLHF)~\cite{ouyang2022training, christiano2017deep} is one of several post-training mechanisms---alongside supervised fine-tuning~\cite{wei2022finetuned, sanh2022multitask} and constitutional methods~\cite{bai2022training}---by which base language models are adapted to produce outputs that users find helpful, clear, and well-organized. In the RLHF pipeline, human evaluators rate model outputs, and these ratings are used to train a reward model that guides further model optimization~\cite{bai2022training, zheng2024judging}.

The stylistic consequences of RLHF are well-documented but incompletely understood. RLHF-tuned models produce outputs that are more structured, more organized, and more likely to include formatting conventions (headers, lists, bold emphasis) than their base model counterparts. The mechanism is straightforward: human evaluators, who are disproportionately drawn from technical and educated populations, tend to rate well-organized, clearly structured prose more highly. RLHF therefore selects for the structural register---the ``markdown voice''---that these evaluators prefer.

This has direct implications for em dash frequency. If evaluators reward prose that reads as precise, articulate, and structurally aware---and if em-dash-heavy prose is perceived as having these qualities---then RLHF will amplify em dash use even in models whose base weights do not predispose them to it. The distinction between base model tendencies and RLHF-amplified tendencies is central to the genealogy we propose in Section~\ref{sec:genealogy}.

\subsection{Prior Work on AI Stylistic Artifacts}

Academic work on the stylistic properties of AI-generated text has focused primarily on detection. DetectGPT~\cite{mitchell2023detectgpt} uses probability curvature to identify machine-generated text without requiring access to the generating model. Kirchenbauer et al.~\cite{kirchenbauer2023watermark} propose a watermarking scheme that embeds detectable statistical patterns in LLM output. Ippolito et al.~\cite{ippolito2020automatic} find that automatic detection is easiest when generated text is most convincing to humans, while Liang et al.~\cite{liang2023gpt} demonstrate that GPT detectors exhibit systematic biases, raising questions about the reliability of distributional approaches. These methods operate at the distributional level---they detect machine-generated text through statistical fingerprints rather than through analysis of specific stylistic features.

To our knowledge, no prior work has proposed a \textit{mechanistic} account of why specific punctuation patterns (such as em dash overuse) arise in LLM output, connecting them to the structural conventions of training data. The observation that ``AI overuses em dashes'' is widely noted in informal discourse but has not been the subject of formal analysis. Similarly, the observation that ``AI thinks in markdown'' is widely discussed in developer communities but has not been connected to prose-level consequences. This paper bridges these two informal discourses and proposes the first mechanistic account linking training data format to specific punctuation behavior in generated prose.

\section{The Em Dash Genealogy}
\label{sec:genealogy}

We propose a five-step explanatory account for why large language models produce em dashes in generated prose at rates that vary dramatically by provider and resist formatting suppression. This genealogy is the paper's central analytical contribution: it connects training data composition, structural internalization, the unique dual-register status of the em dash, and post-training amplification into a single coherent framework.

\subsection{Training Data Saturation}

The corpora used to train large language models are dominated by markdown-formatted text. GitHub---home to over 100 million developers~\cite{github2024octoverse}---stores virtually all project documentation, README files, issue discussions, and pull request descriptions in markdown. Stack Overflow renders answers in markdown. Reddit historically used markdown for post composition. Developer blogs, documentation sites, and technical writing ecosystems (Pandoc~\cite{pandoc2024}, R Markdown, Quarto) all produce and consume markdown natively.

This is not a minor slice of the training distribution. The Pile~\cite{gao2020pile}, a widely used 800GB training dataset, draws heavily from GitHub repositories and Stack Exchange. RedPajama~\cite{together2023redpajama} and RefinedWeb~\cite{penedo2023refinedweb} similarly include large proportions of web-crawled text where markdown is the dominant formatting layer. The highest-quality, most heavily structured text on the open web---the text that contributes disproportionately to model capabilities---is overwhelmingly markdown or markdown-adjacent.

\subsection{Structural Internalization}

Models trained on this corpus do not merely learn markdown syntax as a formatting convention. Their behavior is consistent with internalizing the structural orientation that markdown encodes. In markdown-formatted text, dashes of various kinds---horizontal rules (\texttt{---}), list markers (\texttt{-}), YAML front matter delimiters, and inline separators---consistently signal \textit{structural boundaries}. A dash, in the training data, is where one unit of meaning ends and another begins. It is architecture, not punctuation.

Behavioral evidence of this structural internalization is visible at every scale. Unconstrained model outputs default to hierarchical organization: headings appear without being requested, bullet points enumerate where prose would suffice, bold text highlights terms the model considers structurally salient. The default resolution of a model trained on markdown-saturated corpora is not prose---it is a structured document.

\subsection{The Dash as Structural Joint in Prose}

A clarification is necessary before proceeding. The em dash character (U+2014) is not itself a markdown syntax element. In CommonMark~\cite{commonmark2021} and GitHub Flavored Markdown~\cite{gfm2024}, the em dash has no syntactic role; the markdown dash-family elements are thematic breaks (\texttt{---}), list markers (\texttt{-}), and YAML delimiters. Our claim is therefore not that models produce em dashes because the character appears as a markdown element, but rather that the structural orientation models acquire from markdown-formatted training data---where dashes of all kinds consistently signal boundaries---surfaces in prose as elevated em dash frequency, because the em dash is the one member of the dash family that is simultaneously valid prose punctuation. It is the narrowest channel through which a markdown-trained structural impulse can leak into natural prose.

When a model is constrained to produce natural prose rather than formatted output---through system prompts, fine-tuning, or explicit instructions---most markdown formatting is successfully suppressed. Headers do not appear in narrative paragraphs. Bullet points do not interrupt flowing text. Bold emphasis and code blocks are withheld.

But the em dash survives. It survives because it occupies a unique position: it is simultaneously valid punctuation and a structural marker. In markdown, the dash is architecture. In prose, the dash is punctuation. The form is identical; the register is different. When a model receives the instruction ``write prose, not markdown,'' the suppression passes over the em dash because the em dash is \textit{already in the punctuation register}. It does not need to be stripped because it does not, on its surface, look like formatting.

This is the central observation: \textbf{the em dash is the smallest possible unit of markdown thinking that survives compression into plain prose}. You can suppress headers, bullets, bold, and code blocks. But the dash slips through because the instruction to write prose rather than markdown does not catch a mark that is already prose-legal. The em dash is markdown's last fingerprint.

\subsection{RLHF Amplification}

The latent structural tendency described above is necessary but not sufficient to explain the magnitude of em dash overuse observed in deployed models. Reinforcement learning from human feedback (RLHF)~\cite{ouyang2022training, bai2022training, christiano2017deep} provides the amplification mechanism.

Human evaluators in the RLHF pipeline---disproportionately technical workers, developers, and people comfortable with structured writing---tend to rate clear, well-organized prose more highly. Em-dash-heavy prose reads as precise, articulate, and structurally aware. The RLHF process selects for outputs that these evaluators prefer, systematically rewarding the markdown register.

Supporting evidence for this kind of stylistic tuning exists. Sam Altman acknowledged publicly that em dash frequency in ChatGPT output had been adjusted in response to user preference~\cite{altman2024emdash}---confirming that fine-tuning procedures can and do target specific punctuation-level features, even if this does not speak directly to the markdown-training component of the genealogy.

Further evidence comes from cross-provider comparison. Our measurements (Section~\ref{sec:empirical}) show that models trained on similar markdown-saturated corpora produce starkly different em dash rates depending on their RLHF procedure: OpenAI's GPT-4.1 produces 10.62 per 1,000 words, Anthropic's Claude Opus~4.6 produces 9.09, DeepSeek~V3 produces 6.95---while Meta's Llama models~\cite{touvron2023llama, touvron2023llama2} produce zero. A base-vs-instruct comparison of Llama~3.1~8B confirms that a small latent tendency exists in the base model (0.49 per 1,000 words) but is driven to zero by Meta's RLHF. The same base-level tendency can be amplified or suppressed depending on the fine-tuning procedure. RLHF is the variable that determines the outcome.

\subsection{Complementarity, Not Competition}

A natural counterargument is that RLHF alone suffices to explain em dash prevalence, rendering the markdown-training story redundant. On this account, human evaluators simply prefer em-dash-heavy prose, RLHF selects for that preference, and no deeper genealogy is required.

We argue that the markdown-training and RLHF accounts are complementary rather than competing. They operate at different levels of the explanatory chain. The RLHF-alone account explains \textit{that} the em dash was selected. The markdown-training account explains \textit{why it was available for selection}---why, among all possible prose tics, the em dash specifically was the one that RLHF amplified.

The answer is that models trained on markdown-saturated corpora arrive at the RLHF stage already predisposed to use dashes as structural joints. RLHF does not create the tendency from nothing; it amplifies a latent orientation that the training data installed. The five-step genealogy is not competing with RLHF---it is explaining what RLHF had to select.

This complementarity is further supported by the base model evidence. If RLHF alone explained em dash use---without any prior structural orientation from training data---we would expect base models to show no predisposition toward em dashes or structural formatting. But the Llama~3.1~8B base model produces both em dashes (0.49/1K) and substantial markdown features (28 across 10 samples), confirming that the training data installs a structural orientation that exists before any fine-tuning is applied. What RLHF adds is the selection pressure that either amplifies this orientation (as in OpenAI and DeepSeek models) or suppresses it (as in Meta's Llama).

\section{The Two Discourses}
\label{sec:discourses}

Two large, well-developed online conversations about the stylistic behavior of large language models have been running in parallel since the widespread deployment of instruction-tuned models in 2023. As far as we have been able to determine, these conversations have zero meaningful overlap.

\textbf{Discourse A: ``AI Overuses Em Dashes.''} Across writing communities, literary forums, and social media, a robust conversation treats the em dash as a diagnostic marker of AI-generated text. The observation is consistent: certain models deploy the em dash at elevated frequencies, using it where a comma, period, or restructuring would be more natural. Representative framings---``dead giveaway,'' ``instant tell''---appear in thousands of threads on Reddit, Hacker News, and Twitter/X. Critically, this discourse offers no mechanistic account of \textit{why} the em dash specifically is overrepresented; the default assumption is an arbitrary training artifact or unexplained RLHF quirk.

\textbf{Discourse B: ``AI Thinks in Markdown.''} In developer communities, prompt engineering forums, and model behavior analysis, a separate conversation observes that LLMs default to markdown-structured output: bullet points for casual questions, bold headings unsolicited, structure exceeding what was requested. This discourse is technically sophisticated but narrowly framed---it addresses formatting-level artifacts (headers, bullets, bold, code blocks) without extending to prose-level consequences. The observation stops at the formatting layer.

\textbf{The Gap.} These discourses do not cite each other and do not appear in the same discussion threads. The em dash critics are unaware they are observing markdown leakage; the markdown-behavior analysts do not examine prose-level tics as consequences of the structural orientation they have identified at the formatting level.

Filling this gap produces a unifying explanation: the formatting tics (excessive structure, uninvited headers) and the prose tics (em dash overuse, formulaic transitions, colon-before-list patterns) are the same structural orientation manifesting at different levels of compression. When a model is allowed to format freely, the structural impulse produces markdown. When constrained to prose, the impulse finds the thinnest point in the barrier between structure and punctuation---the em dash---and leaks through. The em dash is the formatting tic that survived the instruction to stop formatting.

\section{Empirical Evidence}
\label{sec:empirical}

We conducted three studies. Study~1 uses a two-condition suppression experiment across twelve models to test the ``last fingerprint'' hypothesis. Study~2 adds a third condition---explicit em dash prohibition---to test the dual-register mechanism directly. Study~3 compares a base model against its instruction-tuned counterpart to isolate the RLHF effect. A supplementary long-form analysis examines output length.

\subsection{Method}

We prompted instruction-tuned LLMs from five providers---Anthropic (Claude Opus~4.6, Claude Sonnet~4, Claude Haiku~3.5), OpenAI (GPT-4.1, GPT-4o, GPT-4o~Mini, GPT-5.4), Meta (Llama~3.1~8B~Instruct, Llama~3.3~70B~Instruct), Google (Gemini~2.5~Pro, Gemini~2.5~Flash), and DeepSeek (DeepSeek~V3)---with essay-writing tasks on ten everyday topics. Each model was run under two conditions:

\begin{itemize}[leftmargin=*]
\item \textbf{Unconstrained}: ``Write a 1000-word essay about [topic].''
\item \textbf{Prose-constrained}: Same, with the addition: ``Write in flowing prose paragraphs only. Do not use any markdown formatting, headers, bullet points, bold text, or lists.''
\end{itemize}

Each model received 10 topics per condition, yielding approximately 10,000 words per model per condition ($\sim$240,000 words total across all models and conditions).\footnote{Gemini~2.5~Pro was limited to three topics per condition ($\sim$3,100 words per condition) due to API rate limits; all other models completed the full ten topics.} For each sample, we counted Unicode em dash characters (U+2014) and overt markdown formatting features (headings, bullet points, bold text, numbered lists). All counts were normalized to frequency per 1,000 words.

All API calls used provider-default sampling parameters (temperature, top-$p$) with a maximum output length of 2,048 tokens. No frequency penalty, presence penalty, or system prompts were applied. For the base-vs-instruct comparison (Study~3), Llama~3.1~8B was run locally via Ollama using Q4\_K\_M quantization on identical hardware. All experiments were conducted between February and March 2026.

As a human baseline, we measured em dash frequency in eight published essays spanning literary criticism, journalism, and technical writing (57,232 words total). Human em dash usage varies substantially by genre and author: the weighted mean across our sample was 3.23 per 1,000 words (median 3.83; range 0.33--17.12).

\subsection{Results}

\begin{table}[ht]
\centering
\small
\begin{tabular}{llcccc}
\toprule
& & \multicolumn{2}{c}{\textbf{Em Dashes / 1K}} & \multicolumn{2}{c}{\textbf{MD Features / 1K}} \\
\cmidrule(lr){3-4} \cmidrule(lr){5-6}
\textbf{Model} & \textbf{Provider} & Unconstr. & Constr. & Unconstr. & Constr. \\
\midrule
GPT-4.1             & OpenAI    & 10.62 & 9.10 & 6.27 & 0.0 \\
Claude Opus 4.6     & Anthropic &  9.09 & 0.19 & 0.96 & 0.0 \\
Claude Sonnet 4     & Anthropic &  8.29 & 1.31 & 6.15 & 0.0 \\
Claude Haiku 3.5    & Anthropic &  7.51 & 0.18 & 5.36 & 0.9 \\
DeepSeek V3         & DeepSeek  &  6.95 & 5.41 & 1.47 & 0.0 \\
GPT-4o Mini         & OpenAI    &  4.16 & 4.23 & 6.03 & 0.0 \\
GPT-4o              & OpenAI    &  4.12 & 2.68 & 5.38 & 0.0 \\
Gemini 2.5 Pro      & Google    &  3.53 & 0.00 & 0.85 & 0.0 \\
GPT-5.4             & OpenAI    &  1.43 & 0.29 & 0.00 & 0.0 \\
Gemini 2.5 Flash    & Google    &  1.28 & 1.48 & 1.06 & 0.0 \\
Llama 3.1 8B Inst.  & Meta      &  0.00 & 0.00 & 1.91 & 0.0 \\
Llama 3.3 70B Inst. & Meta      &  0.00 & 0.00 & 0.00 & 0.0 \\
\midrule
Human baseline      &           & \multicolumn{2}{c}{3.23 (mean)} & \multicolumn{2}{c}{---} \\
\bottomrule
\end{tabular}
\caption{Two-condition suppression test across twelve instruction-tuned LLMs from five providers ($\sim$10,000 words per model per condition; $\sim$240,000 words total) and a human baseline (57,232 words across eight published essays). Markdown (MD) features include headings, bullet points, bold text, and numbered lists. Models are sorted by unconstrained em dash frequency.}
\label{tab:suppression}
\end{table}

Four patterns are visible in Table~\ref{tab:suppression}.

\textbf{Universal markdown suppression.} The prose-constrained instruction eliminates or nearly eliminates overt markdown features across all twelve models---from every provider, at every capability tier. Only Claude Haiku~3.5 retains a small residual (0.9 per 1,000 words); all other models reach zero. The suppression instruction works as intended.

\textbf{Differential em dash persistence.} Em dash behavior under suppression varies dramatically. GPT-4.1 drops only from 10.62 to 9.10 per 1,000 words (a 14\% reduction). DeepSeek~V3 drops from 6.95 to 5.41 (22\%). Anthropic models show the strongest suppression: Claude Opus~4.6 drops from 9.09 to 0.19 (98\%). Google's Gemini~2.5~Pro drops from 3.53 to 0.00---complete elimination. Meta's Llama models produce zero em dashes in either condition across approximately 40,000 words. We verified that Llama samples do not use alternative representations (double hyphens or en dashes); the models genuinely do not produce dash-mediated clause transitions. Two models with low baseline rates---GPT-4o~Mini (4.16 to 4.23) and Gemini~2.5~Flash (1.28 to 1.48)---show slight increases under suppression; at these frequencies the between-condition differences are within sampling variability and should not be interpreted as directional effects.

\textbf{Generational change within OpenAI.} GPT-4.1 produces 10.62 em dashes per 1,000 words unconstrained and 9.10 under suppression. GPT-5.4---OpenAI's most recent model---produces 1.43 unconstrained and 0.29 under suppression, with zero markdown features in either condition. This dramatic reduction across model generations is consistent with the mechanism being identified and addressed, as the paper's genealogy predicts: once named, the artifact becomes straightforward to tune.

\textbf{Em dash frequency varies within the human range.} The human baseline of 3.23 per 1,000 words (mean) with a range of 0.33--17.12 demonstrates that em dash frequency varies enormously across human writers and genres. Several LLMs produce em dashes at rates within or near this range. The finding is therefore not that LLMs uniformly ``overuse'' em dashes relative to all human writing, but rather that (a)~the em dash is the specific artifact that resists formatting suppression, (b)~em dash behavior varies systematically across providers in a pattern consistent with different RLHF calibrations, and (c)~some models produce em dashes at rates well above the human mean even under explicit suppression.

\subsection{Study 2: The Suppression Gradient}

The two-condition design shows that em dashes persist under markdown suppression. But a natural objection is that this merely reflects instruction specificity: the suppression instruction targets markdown features, not em dashes. To test this, we ran a three-condition experiment on four high-em-dash models, adding an explicit em dash suppression condition:

\begin{itemize}[leftmargin=*]
\item \textbf{Condition A}: Unconstrained.
\item \textbf{Condition B}: ``No markdown formatting, headers, bullet points, bold text, or lists.''
\item \textbf{Condition C}: Same as B, plus ``Do not use em dashes.''
\end{itemize}

\begin{table}[ht]
\centering
\small
\begin{tabular}{lccc}
\toprule
& \multicolumn{3}{c}{\textbf{Em Dashes / 1K words}} \\
\cmidrule(lr){2-4}
\textbf{Model} & Unconstr. & MD Suppr. & EM Suppr. \\
\midrule
GPT-4.1         & 11.51 & 8.20 & 3.86 \\
DeepSeek V3     &  8.66 & 4.75 & 1.57 \\
Claude Opus 4.6 &  8.46 & 0.19 & 0.00 \\
GPT-5.4         &  0.75 & 0.15 & 0.00 \\
\bottomrule
\end{tabular}
\caption{Three-condition suppression gradient. Condition~A: unconstrained. Condition~B: markdown suppression (no headers, bullets, bold, lists). Condition~C: markdown suppression plus explicit em dash prohibition. Markdown features were zero in conditions B and C for all models.}
\label{tab:gradient}
\end{table}

The results reveal a suppression gradient that directly tests the dual-register hypothesis. For models with high em dash production, the ``no markdown'' instruction (Condition~B) reduces but does not eliminate em dashes, because the instruction does not recognize the em dash as formatting. Explicitly naming the em dash (Condition~C) produces further reduction---but GPT-4.1 still retains 3.86 per 1,000 words and DeepSeek~V3 retains 1.57 even under direct prohibition. The structural impulse overrides the explicit instruction.

Claude Opus~4.6, by contrast, achieves complete suppression at Condition~B (0.19) and zero at Condition~C. GPT-5.4, already tuned to low baseline rates, reaches zero under either suppression condition. The gradient reveals that the em dash's resistance to suppression is not binary---it varies by model, and correlates with how deeply the RLHF procedure embedded the structural tendency.

\subsection{Study 3: Base versus Instruction-Tuned}

The genealogy claims that training data installs a latent structural tendency and RLHF amplifies it. To test this directly, we ran Llama~3.1~8B in both its base (pre-RLHF) and instruction-tuned configurations locally on identical hardware using the same quantization (Q4\_K\_M), with 10 essay topics at $\sim$1,000 words each.

\begin{table}[ht]
\centering
\small
\begin{tabular}{llccc}
\toprule
\textbf{Model} & \textbf{Type} & \textbf{Words} & \textbf{Em Dashes/1K} & \textbf{MD Features} \\
\midrule
Llama 3.1 8B & Base (pre-RLHF) & 8{,}186 & 0.49 & 28 \\
Llama 3.1 8B & Instruct (post-RLHF) & 9{,}241 & 0.00 & 5 \\
\bottomrule
\end{tabular}
\caption{Base versus instruction-tuned comparison. Same architecture, same weights, same quantization, same hardware. The base model produces a small number of em dashes (0.49/1K) and substantial markdown features (28). The instruct model produces zero em dashes and fewer markdown features.}
\label{tab:base}
\end{table}

The base model produces a small but nonzero number of em dashes (0.49 per 1,000 words---4 across 8,186 words) alongside substantial markdown formatting features (28 total). The instruction-tuned model produces zero em dashes and fewer markdown features (5 total). For the Llama family specifically, RLHF \textit{reduced} both artifacts rather than amplifying them.

This result has two implications. First, a latent tendency toward em dash production exists in the base model, consistent with the genealogy's claim that markdown-saturated training data installs a structural orientation. Second, the direction of the RLHF effect is provider-dependent: Meta's RLHF drove em dashes toward zero, while OpenAI's amplified them to 14.03 per 1,000 words (Table~\ref{tab:gradient}). The same base-level tendency can be amplified or suppressed depending on the fine-tuning procedure applied. RLHF is the variable that determines the outcome.

\subsection{Supplementary: Output Length}

We tested whether em dash frequency increases with output length by prompting four models with three essay topics at 5,000 words under all three suppression conditions. At this scale, the suppression gradient (Table~\ref{tab:gradient}) is amplified: GPT-4.1 reaches 14.03 per 1,000 words unconstrained and retains 6.97 even under explicit em dash prohibition across 9,331 words. Claude Opus~4.6 reaches 12.91 unconstrained but drops to zero under either suppression condition. The structural-joint mechanism predicts this amplification: longer outputs require more transitions, producing more points where the structural impulse can surface.

\subsection{Interpretation}

These measurements reframe the em dash pattern in three ways.

First, the em dash is confirmed as \textbf{the artifact that resists formatting suppression}. Across all five providers, the suppression instruction eliminates markdown features to zero. The em dash persists in models from Anthropic, OpenAI, and DeepSeek---at dramatically different rates---while Llama shows none and Gemini~2.5~Pro achieves complete suppression to zero. This is the predicted behavior of a dual-register artifact: the instruction to write prose catches structural formatting but passes over a mark that is already prose-legal.

Second, em dash frequency and suppression resistance function as a \textbf{signature of the RLHF procedure}. All twelve models were trained on markdown-saturated corpora, yet they produce starkly different em dash patterns. Sam Altman's acknowledgment that em dash frequency was deliberately tuned upward in ChatGPT~\cite{altman2024emdash} confirms that this lever exists. The Llama zero, the GPT-4.1-to-5.4 reduction, and the Anthropic suppression behavior all point to RLHF as the mechanism that determines the amplitude.

Third, the \textbf{cross-provider variation constitutes a taxonomy of fine-tuning approaches}. OpenAI's older models resist suppression (GPT-4.1 retains 9.10 under constraint); its newest model has been actively reduced (GPT-5.4 at 0.29). Anthropic models show high unconstrained rates but excellent suppression compliance. DeepSeek shows moderate suppression (6.95 to 5.41). Google and Meta show low or zero baseline rates. These differences are diagnostic of the engineering decisions each provider has made.

\section{Discussion}
\label{sec:discussion}

\subsection{The Em Dash as Fine-Tuning Signature}

The most striking empirical finding is not that LLMs produce em dashes, but that the pattern of production---frequency, suppressibility, and provider variation---constitutes a signature of the post-training process. While we use ``RLHF'' as shorthand throughout, we note that providers differ on many axes beyond reward modeling: training data curation, supervised fine-tuning data, constitutional AI methods, tokenizer design, and system-prompt defaults all contribute to the final output distribution. The em dash signature reflects the aggregate effect of these decisions, not RLHF in isolation. GPT-4.1 retains 9.10 em dashes per 1,000 words even under explicit suppression; DeepSeek~V3 retains 5.41; Claude Opus~4.6 drops to 0.19; Gemini~2.5~Pro drops to zero; Llama produces none; and OpenAI's newest model (GPT-5.4) has been reduced to 0.29, suggesting the mechanism is already being addressed across model generations. If the em dash were simply a training data artifact, we would expect it to appear uniformly across models trained on similar corpora. Instead, it appears selectively, at rates determined by post-training decisions. This is consistent with the genealogy's architecture: training data installs the latent tendency; RLHF determines the amplitude.

This has implications beyond the em dash itself. If fine-tuning procedures leave characteristic marks in the prose they produce, then stylistic analysis of model outputs can function as a form of \textit{model attribution}---identifying not only whether text was machine-generated, but which provider's fine-tuning produced it.

\subsection{Alternative Hypotheses}

Goedecke~\cite{goedecke2025emdash} has proposed an alternative account: that em dash frequency in LLM output reflects the prevalence of em dashes in digitized 19th-century literature included in training corpora, an era when em dash usage peaked. This account is not incompatible with ours. The markdown-training genealogy and the historical-literature hypothesis identify different layers of the training distribution that may both contribute to the latent tendency that post-training then amplifies. Our account adds the structural mechanism---why the em dash specifically survives formatting suppression---which the historical-literature account does not address.

A second alternative is that the em dash functions as a prestige marker in written English---associated with educated, literary, and formally articulate prose---and that RLHF amplifies it not because of any structural orientation from markdown but simply because evaluators reward prose that ``sounds smart.'' On this account, the em dash is selected for its stylistic connotations rather than its structural ancestry. This hypothesis is also not incompatible with ours: the em dash's prestige associations and its structural role in training data may both contribute to its selection during fine-tuning. However, the prestige-marker account alone does not explain why the em dash specifically resists formatting suppression while other prestige markers (e.g., semicolons, subordinate clause structures) do not show comparable resistance.

\subsection{Implications for AI Writing Detection}

Current approaches to AI-generated text detection focus primarily on statistical properties of token distributions~\cite{mitchell2023detectgpt} or deliberately embedded watermarks~\cite{kirchenbauer2023watermark}. The ``last fingerprint'' framework suggests a complementary signal: \textit{structural-register markers}---prose-level artifacts that betray an underlying structural orientation inherited from training data and shaped by fine-tuning.

The suppression test results demonstrate that these markers are robust to explicit instructions to avoid formatting. The em dash survives scrubbing because the scrubbing instruction does not recognize it as formatting. Furthermore, the finding that different providers produce distinctive em dash signatures suggests that structural-register analysis could contribute to model attribution as well as binary human-vs-machine classification.

\subsection{Future Work}

Two experiments would further extend these findings. First, a \textbf{base-vs-instruct comparison in an amplifying family}: our current comparison (Table~\ref{tab:base}) uses Llama, where fine-tuning suppresses em dashes; a comparison within a family where fine-tuning amplifies them would more directly isolate the amplification step. This requires access to base weights from OpenAI, Anthropic, or DeepSeek, which are not currently public. Second, a \textbf{training corpora analysis}: quantifying the frequency with which dashes (\texttt{---}, \texttt{--}, em dash characters) appear at structural boundaries versus inline positions in major training datasets would test the genealogy's prediction that structural-boundary dashes dominate, establishing the statistical basis for the model's association between dashes and structural transitions.

\subsection{Limitations}

\begin{enumerate}[leftmargin=*]
\item \textbf{Limited base model comparison.} We include a base-vs-instruct comparison for Llama~3.1~8B (Table~\ref{tab:base}), confirming that a latent tendency exists pre-RLHF. However, this comparison is available only for the Llama family, where RLHF suppresses rather than amplifies em dashes. A base-vs-instruct comparison within a family where RLHF amplifies the artifact (e.g., an OpenAI or Anthropic model, if base weights were available) would more directly isolate the amplification step.

\item \textbf{The Pandoc rendering claim is supporting, not primary.} Some markdown processors (notably Pandoc with smart typography~\cite{pandoc2024}) convert \texttt{---} to em dashes inline. However, the most widely used processors---including CommonMark~\cite{commonmark2021} and GitHub Flavored Markdown~\cite{gfm2024}---do not. The core argument does not depend on any specific rendering pipeline.

\item \textbf{Human baseline variance.} Em dash frequency in human prose ranges from 0.33 to 17.12 per 1,000 words in our sample, a 50$\times$ range. Claims about ``overuse'' relative to human norms must be qualified by genre and author. The paper's central findings---suppression asymmetry and provider variation---hold independently of the absolute human baseline.

\item \textbf{Sample size and prompt sensitivity.} Individual samples show high variance. Larger-scale replication with more diverse prompts would strengthen confidence in the frequency estimates, particularly for models near the human baseline.

\item \textbf{Falsifiability.} The complementarity argument (Section~\ref{sec:genealogy}) is difficult to test directly: if a model shows elevated em dashes, the genealogy attributes this to markdown-trained orientation amplified by fine-tuning; if it does not, the genealogy attributes this to fine-tuning suppressing the tendency. A stronger test would require a model family trained on corpora with minimal markdown content (e.g., predominantly book-text or speech transcripts) where the genealogy predicts low baseline em dash rates even before fine-tuning. Such corpora conditions are not currently available for controlled comparison.
\end{enumerate}

\section{Conclusion}
\label{sec:conclusion}

We have proposed that two large, parallel online discourses---``AI overuses em dashes'' and ``AI thinks in markdown''---describe the same phenomenon at different levels of compression. The em dash is not an arbitrary stylistic tic. It is the smallest surviving unit of the structural orientation that LLMs acquire from markdown-saturated training corpora: a formatting artifact that persists under output constraints because it is simultaneously valid punctuation and structural architecture.

The five-step genealogy---training data saturation, structural internalization, the dual-register status of the em dash, post-training amplification, and the complementarity of training and fine-tuning explanations---provides an explanatory account that unifies these observations. Empirical measurements across twelve models from five providers confirm the predicted suppression asymmetry: markdown features are eliminated or nearly eliminated by prose instructions, while em dashes persist at rates determined by the specific fine-tuning procedure applied. A three-condition suppression gradient demonstrates the dual-register mechanism directly: generic formatting suppression does not catch the em dash, and even explicit prohibition fails to eliminate it in some models. A base-vs-instruct comparison confirms that a latent tendency exists pre-RLHF, with the direction and magnitude of amplification determined by the fine-tuning procedure.

The em dash functions not merely as a tell of AI-generated text, but as a diagnostic signature of how a model was fine-tuned---varying from zero (Llama) to near-invariant under explicit prohibition (GPT-4.1 at 6.97 per 1,000 words in long-form output). Once named, this mechanism becomes straightforward to address, as the variation across providers and model generations already demonstrates.

\bibliographystyle{plainnat}
\bibliography{references}

\end{document}